\pdfoutput=1

\documentclass[11pt]{article}

\usepackage[final]{acl}

\usepackage{times}
\usepackage{latexsym}

\usepackage[T1]{fontenc}

\usepackage[utf8]{inputenc}

\usepackage{microtype}

\usepackage{inconsolata}

\usepackage{graphicx}

\usepackage{tcolorbox}
\usepackage{amsmath}
\usepackage{amsfonts}
\usepackage{booktabs}    
\usepackage{pifont}      
\usepackage{multirow}
\usepackage{colortbl} 
\usepackage{amssymb}
\usepackage{utfsym}
\definecolor{myRGBcolor}{RGB}{230,183,69}
\newcommand{\greencheck}{\textcolor{olive}{\usym{2714}}}

\title{
Med-U1: Incentivizing Unified Medical Reasoning in LLMs \\ via Large-scale Reinforcement Learning
}

\author{
 \textbf{Xiaotian Zhang\textsuperscript{1,3}\thanks{Equal contribution.}},
 \textbf{Yuan Wang\textsuperscript{1,3}\footnotemark[1]},
 \textbf{Zhaopeng Feng\textsuperscript{1,3}\footnotemark[1]},
 \textbf{Ruizhe Chen\textsuperscript{1,3}\footnotemark[1]} \\
 \textbf{Zhijie Zhou}, 
 \textbf{Yan Zhang\textsuperscript{2}},
 \textbf{Hongxia Xu\textsuperscript{1,3}},
 \textbf{Jian Wu\textsuperscript{1,3}},
 \textbf{Zuozhu Liu\textsuperscript{1,3}\thanks{Corresponding author.}}
\\
 \textsuperscript{1}Zhejiang University
 \textsuperscript{2}Bytedance Inc \\
 \textsuperscript{3}Zhejiang Key Laboratory of Medical Imaging Artificial Intelligence
\\
   \{xiaotian.24, yuan2.24, zuozhuliu\}@intl.zju.edu.cn
}

\begin{document}
\maketitle
\begin{abstract}

Medical Question-Answering (QA) encompasses a broad spectrum of tasks, including multiple choice questions (MCQ), open-ended text generation, and complex computational reasoning. Despite this variety, a unified framework for delivering high-quality medical QA has yet to emerge. Although recent progress in reasoning-augmented large language models (LLMs) has shown promise, their ability to achieve comprehensive medical understanding is still largely unexplored. In this paper, we present Med-U1, a unified framework for robust reasoning across medical QA tasks with diverse output formats, ranging from MCQs to complex generation and computation tasks. Med-U1 employs pure large-scale reinforcement learning with mixed rule-based binary reward functions, incorporating a length penalty to manage output verbosity. With multi-objective reward optimization, Med-U1 directs LLMs to produce concise and verifiable reasoning chains. Empirical results reveal that Med-U1 significantly improves performance across multiple challenging Med-QA benchmarks, surpassing even larger specialized and proprietary models. Furthermore, Med-U1 demonstrates robust generalization to out-of-distribution (OOD) tasks. Extensive analysis presents insights into training strategies, reasoning chain length control, and reward design for medical LLMs. Our code is available \href{https://github.com/Monncyann/Med-U1}{here}.

\end{abstract}

\section{Introduction}

\begin{figure}[t]
  \includegraphics[width=0.98\columnwidth]{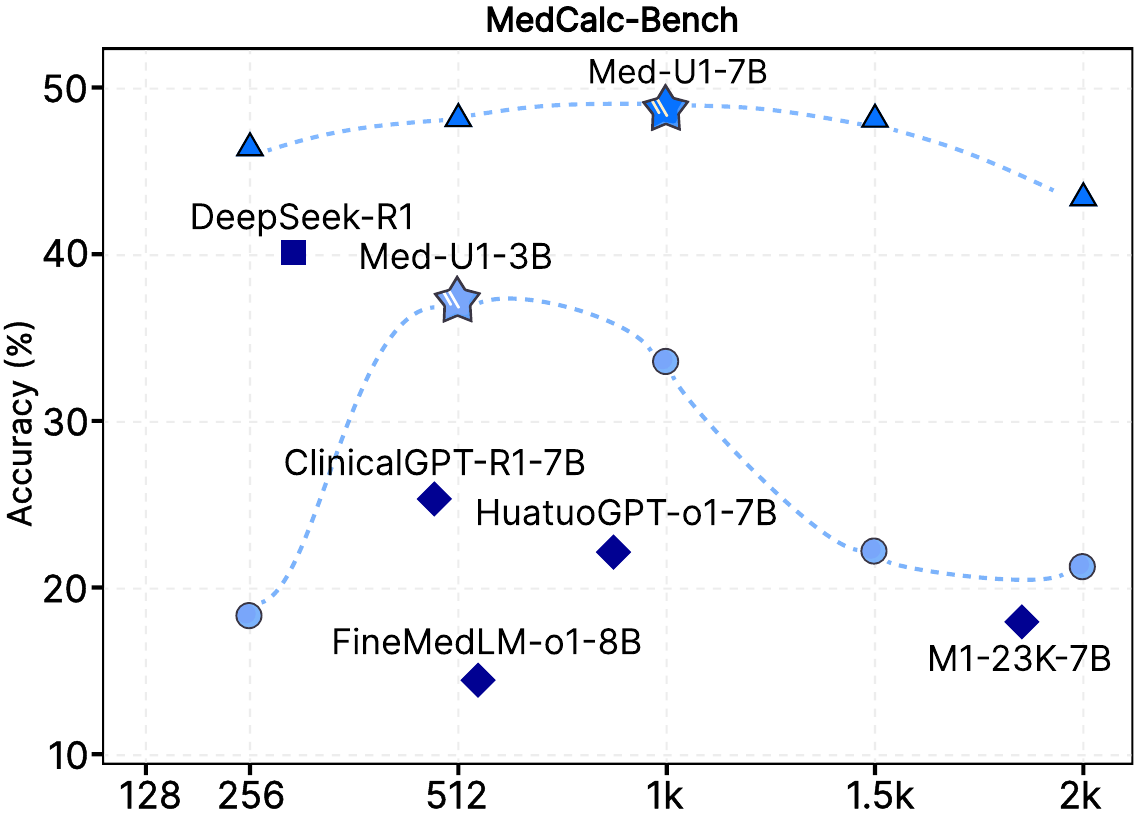}
  \caption{Med-U1 is a length-controllable reasoning model trained with a length-aware reward, achieving superior performance compared to other models. It allows users to specify the desired reasoning length based on task requirements.}
  \label{fig:teaser fig}
\end{figure}

Recent advances in reinforcement learning (RL) based long-form reasoning for medical scenarios have significantly improved model performance of large language models (LLMs) on complex tasks~\citep{zhang2024ultramedical,griot2025large,lievin2024can}. 
Unlike domains such as mathematical reasoning or code generation, where answers are typically verifiable and label formats are relatively consistent~\cite{jaech2024openai,team2025kimi,teamqwq}, medical reasoning spans a wide range of types, including conditional inference, numerical computation, open-ended generation and etc, each exhibiting substantial variation in difficulties and output formats~\citep{medcalc,medxpertqa,chen2024huatuogpt,liu2024medcot}. Beyond standard multiple choice questions (MCQs) with verifiable answers, medical tasks also include computational reasoning for patient care (e.g., hospital stay duration or fluid maintenance) and open-ended text generation with short phrases or long-form text, requiring deep understanding and varied lengths of reasoning, as illustrated in Figures~\ref{fig:dataset} and~\ref{fig:dataset-app}.



Recent research highlights that large-scale RL, such as DeepSeek-R1-Zero~\cite{guo2025deepseek}, can achieve promising performance using binary rule-based reward signals without requiring extensive human-annotated data for Supervised Fine-Tuning (SFT) or Chain-of-Thought (CoT). Several efforts have adapted RL-style training paradigms to the medical domain. \citet{chen2024huatuogpt} utilized reasoning traces generated from MCQ datasets to perform SFT, followed by proximal policy optimization (PPO) for RL. Similarly, \citet{finemedlm} employed multi-stage SFT with synthetic data and long-form reasoning samples, enhancing reasoning capabilities through Direct Preference Optimization (DPO). However, these methods still rely heavily on large-scale supervised data and underexplore pure RL's potential to improve reasoning. More critically, they focus on tasks with verifiable answers, such as MCQs, limiting their applicability to diverse medical reasoning tasks and real-world clinical scenarios (e.g., computation or generation).


Moreover, efficient inference with high accuracy is critical across diverse computational environments, particularly given the varying difficulty and complexity of medical tasks~\citep{huang2025o1}. Recent studies indicate user preferences for controllable reasoning lengths, favoring detailed explanations for complex queries and concise responses for simpler ones~\citep{aggarwal2025l1}. However, existing methods rarely address this need, especially in the medical domain. For instance, \citet{huang2025m1} introduced a "budget forcing" technique to compel continued reasoning. While effective, this approach incurs significant computational overhead. Results in Figure~\ref{fig:teaser fig} suggest that flexible, dynamic control of reasoning length could reduce inference costs and enhance performance, with Med-U1-3B achieving optimal results at 512 tokens. (See Appendix~\ref{appendix:related work} for related work details.)

In this paper, we propose \textbf{Med-U1}, a unified and pure RL framework with controllable-length reasoning across diverse medical QA tasks, ranging from MCQs to complex generation and computation tasks. We propose a mixed medical task-specific reward function, which extends the binary rule-based rewards to effectively guide training in a variety of medical contexts. Specifically, we explore reward optimization for different medical reasoning tasks, including MCQs, numerical value computation, and open-ended text generation. Furthermore, we design a reward to incorporate length constraints  to dynamically regulate output verbosity. This multi-objective reward design enables LLMs to improve test-time performance via verifiable reasoning while maintaining controllability of reasoning length. Experimental results on 5 challenging benchamarks demonstrate that Med-U1 substantially improves model performance across diverse tasks. The Med-U1-7B surpasses larger-scale models such as Qwen-32B, as well as strong proprietary baselines. Furthermore, Med-U1 exhibits strong out-of-distribution (OOD) generalization capabilities, as evidenced by its superior performance on the MMLU-Pro-health benchmark. We further conduct ablation studies on different initialization strategies and reward guidance configurations, revealing that medical tasks exhibit heightened sensitivity to variations in RL paradigms. In addition, we explore the task-dependent optimal inference lengths across diverse scenarios and analyze the effects of different reward signal designs. Our contributions are:

\begin{itemize}
    \item We present a pure RL framework driven by verifiable reward signals, tailored for diverse medical scenarios, achieving strong performance across both in-distribution and OOD tasks, demonstrating its robustness and adaptability in diverse clinical settings.
    \item We propose a length constraint reward to dynamically balance task complexity and reasoning cost, and analyze how different medical tasks exhibit varying preferences and sensitivities to the reasoning length.
    \item Extensive experiments and ablation studies demonstrate that models optimized solely via RL achieve consistent performance gains across various medical tasks. These results underscore the substantial potential of RL as a generalizable optimization paradigm for medical applications.
\end{itemize}

\begin{figure*}[t]
  \includegraphics[width=2.08\columnwidth]{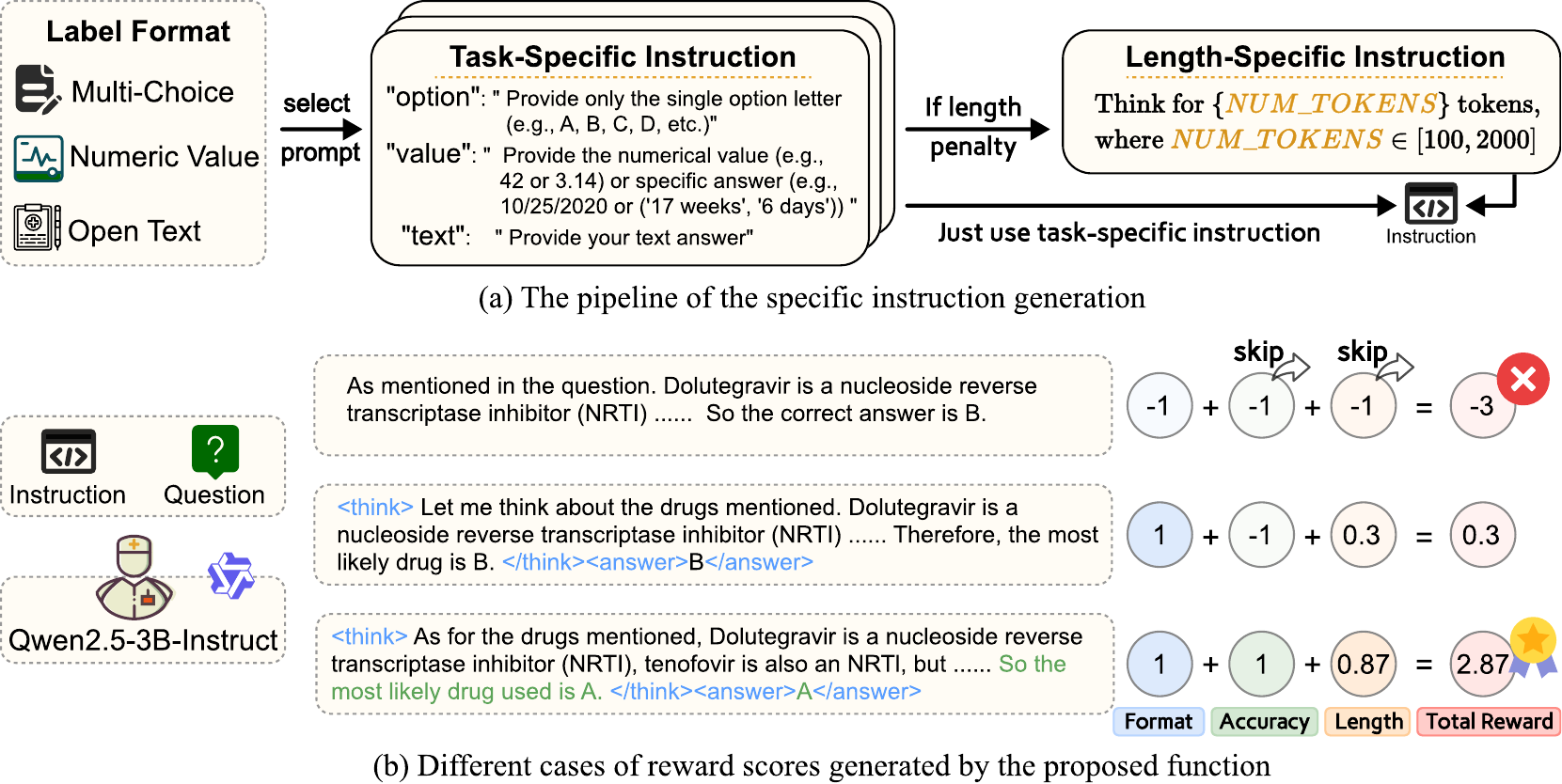}
  \caption{Overall pipeline of generating specific instructions and reward scores. Figure~\ref{fig:pipeline}(a): the generation of instructions based on three output formats. Figure~\ref{fig:pipeline}(b): example of mixed rewards assigned to different responses.}
  \label{fig:pipeline}
\end{figure*}

\section{Method}

The overall pipeline of our method and medical tasks are shown in Figure~\ref{fig:pipeline}. We categorize existing medical reasoning tasks into three types based on output format: multi-choices, numeric value and open text, each might contain further fine-grained formats. Given the generated instructions and questions, our model is trained with pure RL guided by a mixed reward with format validation, correctness verification, and length constraint.


\subsection{Task-specific Reward}
\label{sec:Task-specific Reward}
In RL, the reward signal serves as the primary driver for model optimization. Unlike mathematics and coding, where answers are typically verifiable, medical tasks exhibit diverse output formats. As shown in Figure~\ref{fig:pipeline}, in this paper, we classify these formats into three categories: (1) multiple choice responses with explicit lettered options, (2) numerical values, such as exact numbers short phrases, and (3) open-ended free-text responses. To address this diversity, we propose a task-specific reward function with two components: a format reward that evaluates structural compliance of outputs and a correctness reward that assesses content accuracy.

\noindent \textbf{Format Reward.} Regardless of the task type, the model is required to include its reasoning process within <think></think> tags and its predicted answer within <answer></answer> tags in order to obtain a higher format reward. We employ regular expressions to extract the predicted answer from the output, which is then compared to the ground truth for further correctness evaluation. The format reward $R_{format}$ is computed as:
\begin{equation}
R_{format} =
\begin{cases}
1, & \text{if format is correct,} \\
-1, & \text{if format is incorrect.}
\end{cases}
\end{equation}
\noindent \textbf{Correctness Reward.} 
Only after the response format has been verified do we proceed to assess the quality of the model's prediction. The answer quality score $R_{correct}$ is computed as:
\begin{equation}
R_{correct} =
\begin{cases}
1, & \text{if answer is correct,} \\
-1, & \text{if answer is incorrect.}
\end{cases}
\end{equation}
We explore several task-specific metrics to verify answers according to the nature of the task:
\begin{itemize}
    \item \textbf{MCQs (Options):}
    The evaluation is the same as math benchmarks. The predicted option is extracted from the <answer></answer> tags using regular expressions and directly compared to the ground truth (true o false).
    \item \textbf{Numeric Value Computation:} In ths category, some tasks involve numerical reasoning under specified constraints and output a number (e.g., 42 or 3.14). In other tasks, the prompt provides contextual reasoning cues, and the model is required to output short phrases in specific formats, such as treatment durations (e.g., "17 weeks", "6 days") or dates (e.g., "10/25/2020"). For such tasks with numeric value answers, $R_{correct}$ is determined by verifying that the predicted value lies within the allowable range (that is, $\text{lower limit} \le \text{predicted value} \le \text{upper limit}$), or by performing a direct character-level match against the ground truth.
    \item \textbf{Open-Ended Text Generation:} \label{section: open-ended}
    Open-ended QA represents the most common task in real-world clinical scenarios, for example, generating diagnostic conclusions or treatment plans based on the information provided. These tasks are particularly challenging due to the inherent variability in output formats. To assess the quality of generated content, metrics such as BLEU~\citep{papineni2002bleu} and Rouge-L~\citep{lin2004rouge} are commonly used.
    \begin{tcolorbox}[
    colframe=myRGBcolor,
    colback=yellow!2!white,
    coltitle=white, 
    fonttitle=\bfseries,
    title=Failure Case of BLEU Reward\label{failure case}, 
    boxrule=0.5mm,
]
\textbf{Reference:} At term, the ratio of the weight of the fetus to the weight of the placenta is typically about 6:1.\\
\textbf{Model Output:} 6:01.
\textbf{Bleu:} 0.01.
\label{failure case of bleu}
\end{tcolorbox}
    However, we observe the above faluire case (in yellow box) of BLEU reward in our experiments.
    Specifically, model outputs within the <answer></answer> tags are often concise factual spans, resulting in token-level differences that BLEU fails to reward correctly due to its emphasis on partial n-gram overlap. In cases where the model generates a concise yet accurate description, BLEU may yield misleadingly low scores. To address this issue, we adopt Exact Match Score\footnote{https://huggingface.co/spaces/evaluate-metric/exact\_match} (EMS) as the main reward signal during training. EMS better reflects the correctness of short structured outputs by explicitly measuring word-level match. To provide a softer reward gradient and accommodate near-miss cases, we further incorporate Rouge-L as an auxiliary reward. Rouge-L captures the longest common subsequence between candidate and reference, thus encouraging reward even when full exact matches are not present. In summary, we define a mix evaluation criterion that takes the average of the Rouge-L score and the EMS, assigning equal weight to each component. In addition, we explore training variants that either Rouge-L or EMS is used independently as the sole reward signal. Given that both Rouge-L and EMS produce continuous-valued evaluation scores, we consider an answer correct if its score exceeds a predefined threshold $\tau$.
\end{itemize}
\subsection{Target Length Matching Reward}
Unconstrained reasoning models often suffer from overthinking, where excessive and unnecessary reasoning is performed for simple tasks, leading to computational inefficiency. Conversely, such models may also terminate reasoning prematurely in complex scenarios that require deeper inference. In medical domains with varying levels of task complexity, achieving fast and accurate reasoning with limited computational resources is critical. This requires that models' reasoning length can be dynamically aligned with user-specified requirements.

 Given a question $q$ and a target length $l_{gold}$, the model is encouraged to produce a response $y$ with a total thinking length $l_y$ such that the absolute difference $|l_y - l_{gold}|$ is minimized to maximize the reward $R_{length}$. We define the $R_{length}$ as follows:
 \begin{equation}
 \small
R_{length} = \max\left( -1, \min\left( 1, 1 - \frac{\lvert l_y - l_{\text{gold}} \rvert}{\alpha} \right) \right),
\end{equation}
where $\alpha$ is a normalization factor related to the max output length, and the length reward is ultimately clipped to the interval [-1,1]. By combining the length-based reward with the previously established reward, we encourage the model to generate correct answers through faithful reasoning while simultaneously adhering to user-defined length constraints. This joint reward formulation promotes both correct and efficient reasoning in accordance with user expectations.



\begin{figure*}[t]
  \includegraphics[width=\textwidth]{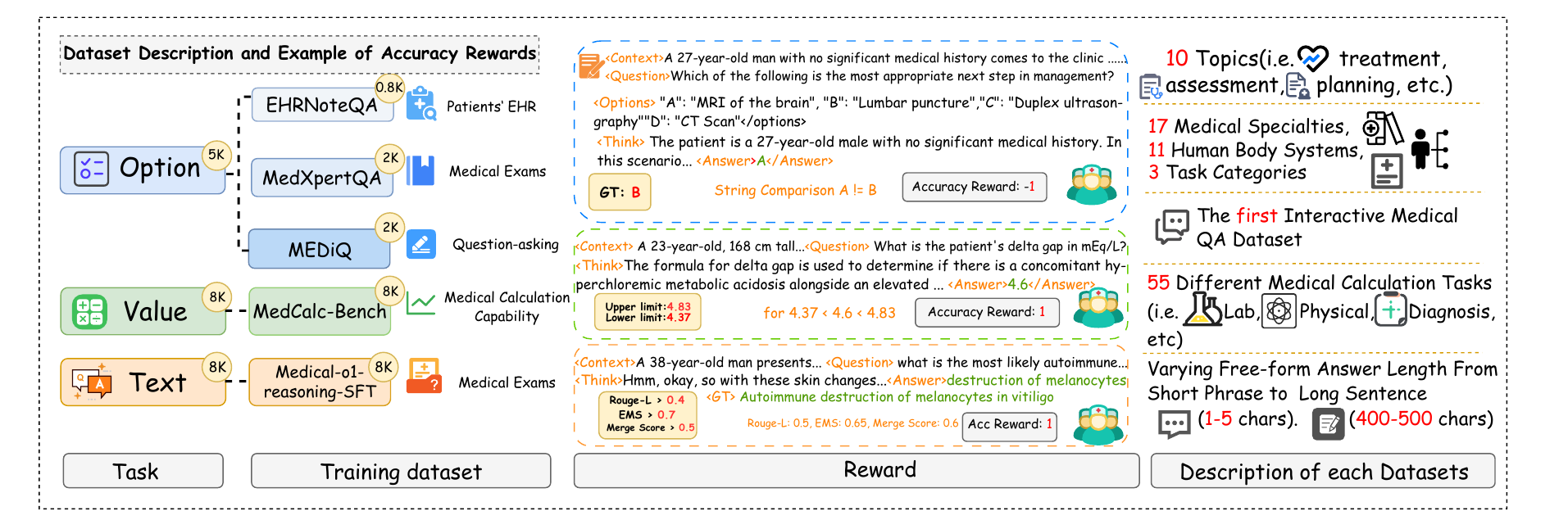}
  \caption{
  Datasets overview for training and in-distribution evaluation, with examples on reward computation.
  }
  \label{fig:dataset}
\end{figure*}

\subsection{Policy Optimization}

As shown in Figure~\ref{fig:pipeline}, the final mixed reward combines three terms: a format reward $R_{format}$, a correctness reward $R_{correct}$ and a target length matching reward $R_{length}$. $R_{length}$ is optional; when omitted, the model is encouraged to reason freely based on its own understanding of the question. The reward function serves a dual purpose: (a) it encourages the model to follow the required format and produce correct answers while implicitly favoring concise reasoning trajectories when shorter outputs are requested by the user; and (b) it consistently incentivizes the model to match the specified target length, thereby facilitating the exploration of optimal reasoning lengths for different task types. 

We use the Group Relative Policy Optimization (GRPO) algorithm to train the model with the final mixed reward. In each training step, for the given question $q$, a group of candidate outputs $O=\{o_1,o_2,\cdots,o_G\}$ are sampled from the policy model $\pi_{\theta_{old}}$. The advantage $A_i = \frac{r_i - \operatorname{mean}(\{r_1, r_2, \dots, r_G\})}{\operatorname{std}(\{r_1, r_2, \dots, r_G\})}$ is calculated using the mixed rewards $\{r_1,r_2,\cdots,r_G\}$. Then the following objective function is maximized to optimize $\pi_{\theta}$:  
\begin{equation}
\begin{aligned}
\small
J_{\mathrm{GRPO}}(\theta) 
&= \mathbb{E}_{q \sim P(Q),\, \{o_i\}_{i=1}^G \sim \pi_{\theta_{\mathrm{old}}}(O \mid q)} \\
&\Biggl[
  \frac{1}{G} \sum_{i=1}^G
  \min\!\Bigl(
    \frac{\pi_{\theta}(o_i \mid q)}{\pi_{\theta_{\mathrm{old}}}(o_i \mid q)}\,A_i,\, \\
    &\mathrm{clip}\!\Bigl(
      \frac{\pi_{\theta}(o_i \mid q)}{\pi_{\theta_{\mathrm{old}}}(o_i \mid q)},
      1-\varepsilon,\,
      1+\varepsilon
    \Bigr)
    A_i
  \Bigr)  \\
  &-\,\beta\,D_{\mathrm{KL}}\bigl(\pi_{\theta}\,\big\|\,\pi_{\mathrm{ref}}\bigr)
\Biggr],
\end{aligned}
\label{eq1}
\end{equation}
where $\varepsilon$ and $\beta$ are hyperparameters controlling the PPO clipping threshold and the weight of the Kullback–Leibler (KL) divergence penalty~\cite{schulman2017proximal,shao2024deepseekmath}, respectively. $\varepsilon$ determines the permissible range for policy updates, while $\beta$ controls the KL divergence penalty.

\section{Experiments}
\subsection{Experimental Settings}
\noindent \textbf{Dataset and Benchmarks.}
\label{section: datasets}
We select medical datasets which are categorized into different task scenarios and cover a variety of label formats: EHRNoteQA~\citep{ehrnoteqa}, MedXpertQA~\citep{medxpertqa}, MEDiQ~\citep{mediq}, MedCalc-Bench~\citep{medcalc}, medical-o1-reasoning-SFT (HuaTuoGPT)~\citep{chen2024huatuogpt}. More details on the above benchmarks are provided in Appendix~\ref{appendix:dataset}. It is worth noting that we excluded intuition-based datasets, as they typically provide limited contextual information in the prompts and are therefore less suitable for supporting long-form reasoning. For the OOD test, we use the health subset of MMLU-Pro~\citep{wang2024mmlu} which consists of 818 challenging multiple choice questions related to healthcare and medicine.

\noindent \textbf{Baselines.} 
Our primary baselines encompass leading proprietary models, namely GPT-4.1, DeepSeek-R1, and Gemini 2.5 Flash, as well as advanced open-source models such as the Qwen2.5 series~\citep{yang2024qwen2} and Llama-3.1-70B-Instruct. We also include strong domain-specific baselines, including HuatuoGPT-o1, FineMedLM-o1, ClinicalGPT-R1 and M1 series, which are specifically tailored for medical applications. Proprietary models are accessed via their APIs. More evaluation details can be found in Appendix~\ref{appendix:evaluation}.

\noindent \textbf{Evaluation Metrics.}
For tasks with available ground truth labels, we employ accuracy as the evaluation metric. In contrast, for open-ended text generation tasks, we evaluate the output quality using a set of complementary metrics: Exact Match Score (EMS) and Rouge-L, which together provide a comprehensive assessment of lexical overlap, and sequence-level correspondence.
\begin{table*}[t]
\centering
\renewcommand{\arraystretch}{1.1}
  \resizebox{0.98\textwidth}{!}{
  \begin{tabular}{l|ccccc|cc}
    \toprule
    \multirow{2}{*}{Baselines} & \multirow{2}{*}{MEDiQ} & \multirow{2}{*}{EHRNoteQA} & \multirow{2}{*}{MedXpertQA} & \multirow{2}{*}{$\text{AVG}_{option}$} & \multirow{2}{*}{MedCalc-Bench}  & \multicolumn{2}{c}{medical-o1-reasoning-SFT} \\
    \cmidrule(lr){7-8}
         & & & & & & Rouge-L & EMS \\
    \hline
    \multicolumn{8}{l}{\cellcolor[HTML]{FFF5D0}\textit{Leading Proprietary Models}}\\
GPT-4.1-nano & 64.84 & 56.16 & 11.02 & 34.60 & 27.53 &  20.78 & 64.41 \\
Gemini-2.5-flash-lite & 85.94 & 56.32 & 31.02 & 51.16 & 38.87 & 24.24  &76.48  \\
DeepSeek-R1 & 78.91 & 63.22 & 20.00 & 44.64 & 40.16 &  27.45  & 41.52  \\
\hline
\multicolumn{8}{l}{\cellcolor[HTML]{FFF5D0}\textit{Medicine-Specific Reasoning LLMs}}\\
Qwen2.5-3B-SFT & 41.41 & 41.38 & 13.06 & 26.33 & 33.90 & 32.42 & 45.55 \\
Qwen2.5-7B-SFT & 64.84& 42.53 & 12.65& 32.89 & 43.34 & 25.32 & 63.44 \\
HuatuoGPT-o1(7B) & 72.66& 44.83 & 14.29 & 36.39 & 22.11  & 14.97 & 21.92 \\
FineMedLM-o1(8B) & 60.16& 40.23 & 10.61& 30.07& 14.31 &  20.00 & 35.58 \\
ClinicalGPT-R1(7B) & 56.25& 43.68 & 11.84 & 30.28 & 18.59 &  22.78 & 32.79 \\
M1-1K(7B) & 60.16 & 54.02& 8.98 & 31.80 & 1.99 & 20.51 & 29.59 \\
M1-23K(7B) & 57.03 & 47.13& 13.06 & 31.79 & 18.69&  25.54 & 39.13 \\
M1-1K(32B) & 67.97 & 58.62 & 18.78 & 40.06 & 19.18 & 22.35 & 32.40 \\
\hline
\multicolumn{8}{l}{\cellcolor[HTML]{FFF5D0}\textit{General Purpose LLMs}}\\
Qwen2.5-3B-Instruct & 53.12& 37.93 & 9.39 & 27.01 & 15.41 & 17.22 & 52.87 \\
Qwen2.5-7B-Instruct & 60.94& 39.08 & 8.16 & 28.76 & 22.66 & 19.98 & 60.56\\
Qwen2.5-32B-Instruct & 70.31 & 44.83 & 8.57 & 32.69 & 33.30& 29.83 & 52.07 \\
Qwen2.5-72B-Instruct & 53.91 & 29.89 &  6.12 & 23.98 & 33.67 & 25.80  & 69.38  \\
Llama-3.1-70B-Instruct &78.91  & 57.47  &18.37  &42.69  &15.85  &21.17  & 57.63  \\
\hline
\multicolumn{8}{l}{\cellcolor[HTML]{FFF5D0}\textit{The proposed}}\\
Med-U1-3B (Ours) & 50.78 & 55.17 & 12.90 & 31.48 & 39.07 &  39.69 & 69.80 \\
Med-U1-7B (Ours) & 65.63 & 52.87 & 17.74& 37.77 & 57.55 &  39.19 & 73.06 \\
\bottomrule
  \end{tabular}}
  \caption{Performance comparison on in-distribution medical tasks. We report average accuracy ($\text{AVG}_{option}$) across 3 MCQ tasks. Open-source baselines are categorized by accessibility and domain specificity. The results are reported in terms of accuracy (\%) and percentage-based scores.}
  \label{tab:in domain}
\end{table*}

\noindent \textbf{Training Details.}
Our implementation is based on the verl\footnote{https://github.com/volcengine/verl} framework. We utilize Qwen2.5-3B/7B-Instruct as the backbone model for training. During training, we configure a batch size of 4 and utilize 8 rollouts per prompt within the GRPO algorithm. For open-ended tasks training, we set the threshold $\tau$ as 40, 50 and 70 for Rouge-L, Mix and EMS reward training, respectively. We employ a constant learning rate of 5e-7 and set the sampling temperature to 1.0. The maximum generation length for responses is capped at 2048 tokens. We set the KL penalty coefficient to 0.001. All models are trained for 1 epoch on 4 NVIDIA A100 80G GPUs.

\begin{table*}[t]
\centering
  \resizebox{0.9\textwidth}{!}{
  \begin{tabular}{l|cccccccc}
    \toprule
 & Nutrition & C-M & C-K & P-M & H-A & Anatomy & M-G & Virology \\
    \midrule
Qwen3B-SFT & 26.26  & 18.75  & 27.78  & 22.05 & 20.93 & 16.46  & 27.78 & 43.48  \\
Med-U1-3B & 43.58 & 47.92 & 45.83 & 37.01 & 33.72 & 24.05 & 53.70 & 36.96 \\
Qwen7B-SFT & 52.17  & 41.67  & 47.22 & 42.91  &30.23  &34.18  & 46.32  &{\cellcolor[rgb]{1,0.925,0.658}}52.17  \\
Med-U1-7B & {\cellcolor[rgb]{1,0.925,0.658}}60.34 &{\cellcolor[rgb]{1,0.925,0.658}} 64.58 & {\cellcolor[rgb]{1,0.925,0.658}}61.11 &{\cellcolor[rgb]{1,0.925,0.658}} 58.66 &{\cellcolor[rgb]{1,0.925,0.658}} 40.70 & {\cellcolor[rgb]{1,0.925,0.658}}48.10 & {\cellcolor[rgb]{1,0.925,0.658}}61.11 & {\cellcolor[rgb]{1,0.925,0.658}}52.17 \\
\bottomrule
  \end{tabular}}
  \caption{Performance comparisons across different medical subdomains, where C-M denotes College Medicine, C-K denotes Clinical Knowledge, P-M denotes Professional Medicine, H-A denotes Human Aging, and M-G denotes Medical Genetics. Results are reported in accuracy (\%).}
  \label{tab: ood}
\end{table*}

\subsection{Main Results}

\noindent \textbf{In-Distribution Performance.}
\label{section: in domain}
Results demonstrate that Med-U1 achieves substantial improvements over its corresponding base versions. Furthermore, it consistently outperforms existing medical reasoning models of comparable scale. It exhibits competitive performance even when compared to significantly larger open-source and proprietary models (Table~\ref{tab:in domain}). On relatively simple benchmarks such as MEDiQ and EHRNoteQA, most baseline models achieve over 40\% accuracy. However, on the more challenging MedXpertQA benchmark, our Med-U1-7B model attains an accuracy of 17.74\%, outperforming nearly all open-source baselines and rivaling 70B-scale general-purpose model and proprietary models. On MedCalc-Bench, which requires higher precision and numerical sensitivity in arithmetic reasoning, the pure RL-based training paradigm significantly enhances the model's inferential and computational abilities, with Med-U1-7B reaching 57.55\% accuracy. In the open-ended generation task, Med-U1 outperforms other methods in both Rouge-L and EMS metrics. We attribute its superior EMS performance primarily to the format reward introduced during training, which encourages the model to place the predicted answer precisely within the <answer></answer> tags, facilitating better alignment with the ground truth. In general, Med-U1 shows leading performance not only in answer matching but also in information coverage and semantic consistency.

\begin{figure}[t]
  \includegraphics[width=0.98\columnwidth]{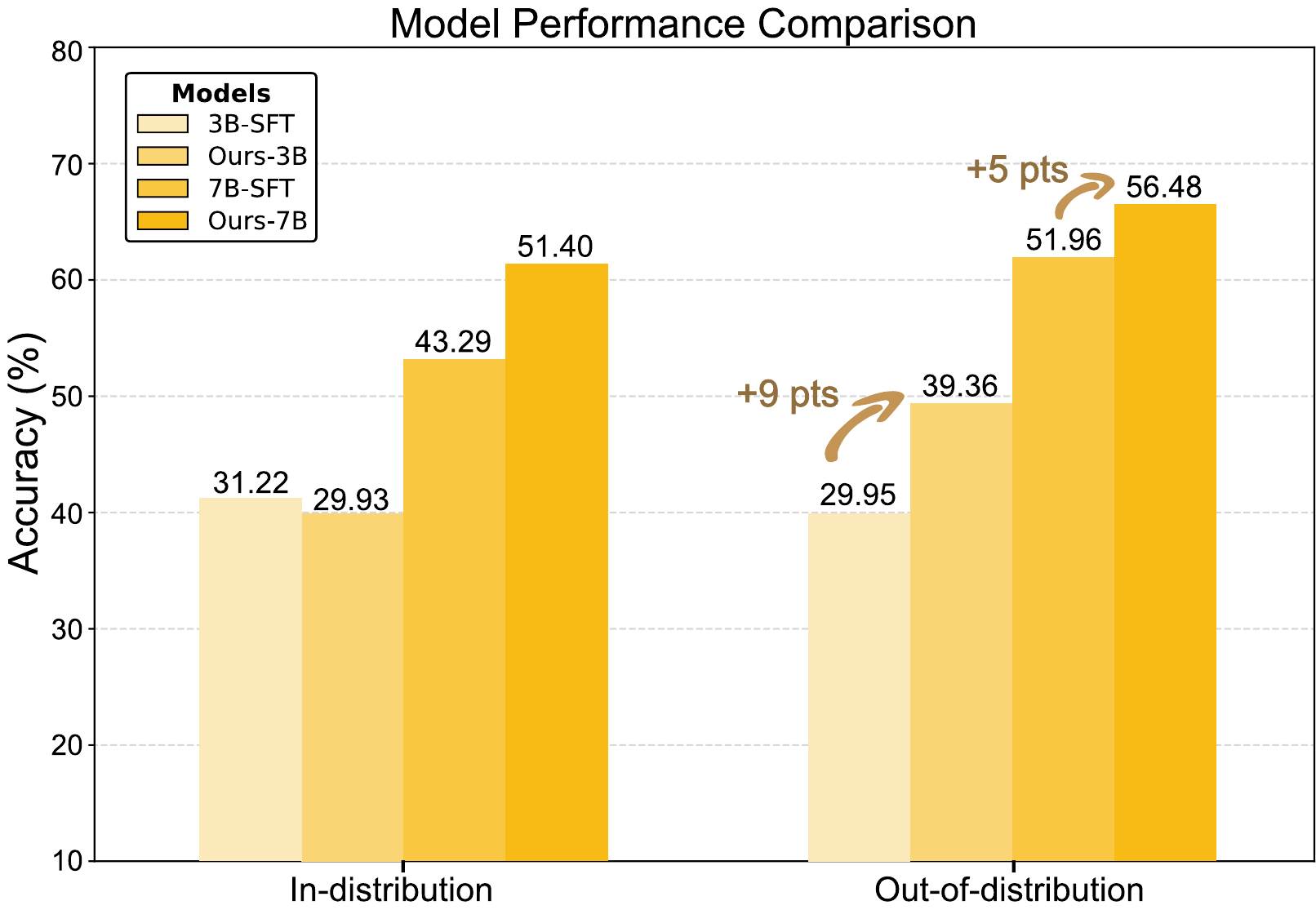}
  \caption{Comparing the proposed method and SFT on in-distribution and out-of-distribution tasks.}
  \label{fig:ood}
\end{figure}

\noindent \textbf{Out-of-Distribution Performance.}
As shown in Table~\ref{tab: ood}, we evaluate eight OOD medical scenarios, with the overall results presented in Figure~\ref{fig:ood}. RL configurations (Med-U1) outperform the SFT baseline, particularly in OOD settings, with the improvements on 3B model can reach up to 9 percentage points. The finding suggests that RL improves generalization, which has been observed in the general domain, also holds in medical tasks. While the use of <think></think> tags may lead the model to reason in reliable patterns, we argue that the primary performance gains stem from the RL paradigm itself, especially given the inherent randomness in OOD data distributions. Several scenarios such as College Medicine and Medical Genetics, where the input often contains richer context to support reasoning, Med-U1-7B shows a substantial performance gain of approximately 19 percentage points compared to the SFT baseline. This aligns with the intuition behind online reinforcement learning: by evaluating and iteratively improving upon self-generated outputs, the model primarily learns how to solve rather than merely how to fit annotated data distribution.

\section{Findings and Analysis}

\begin{figure}[t]
  \includegraphics[width=0.98\columnwidth]{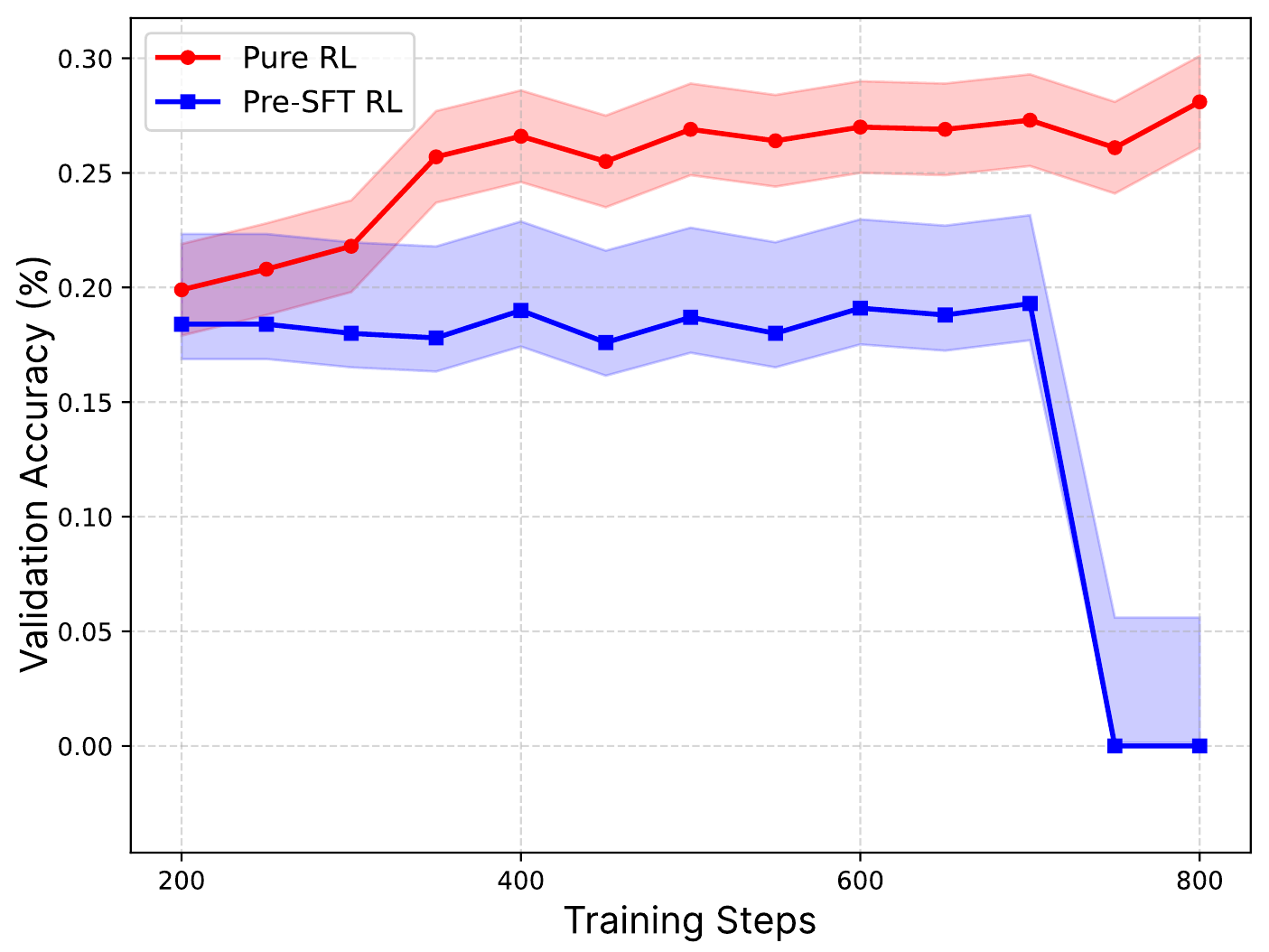}
  \caption{Training dynamics under different training strategies. The red curve denotes the pure RL paradigm, while the blue curve represents the SFT-then-RL approach.}
  \label{fig:medcalc finding}
\end{figure}

\noindent \textbf{Task Sensitivity.}
\label{section: task sensitivity}
As shown in Figure~\ref{fig:medcalc finding}, training with pure reinforcement learning in the MedCalc dataset leads to a more stable validation accuracy and improved performance relative to the pre-SFT-then-RL paradigm. Since MedCalc-Bench involves numerical reasoning tasks in a medical context, it requires models to possess strong computational reasoning abilities. Although SFT enriches the model with medical domain knowledge, it may interfere with the underlying general reasoning capabilities~\citep{huang2025vision,abdin2025phi}. And as provided in Appendix, Figure~\ref{fig:training dynamic} (c) and (d) further validate this trend: SFT provides a strong initialization by enhancing the model’s understanding of domain-specific terms, whereas training solely with RL leads to more stable and gradually improving performance.

\begin{figure}[t]
  \includegraphics[width=0.98\columnwidth]{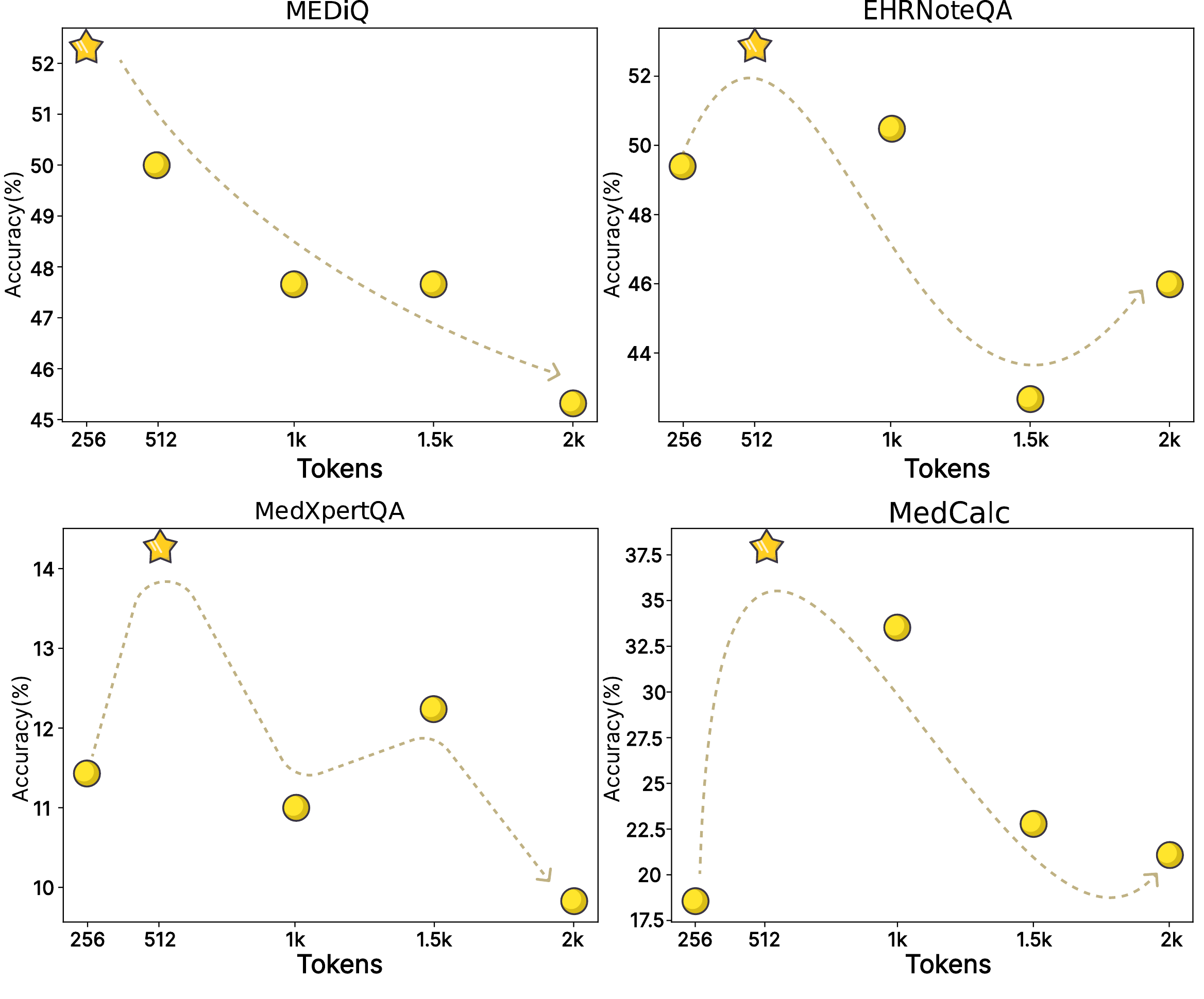}
  \caption{Accuracy of the Med-U1-3B output on different datasets when constrained to specific token lengths during inference. The trend (yellow dashed line) reflects how reasoning length affects task accuracy.}
  \label{fig:token_acc}
\end{figure}

\begin{table}
\centering
\small
\setlength{\extrarowheight}{0pt}
\addtolength{\extrarowheight}{\aboverulesep}
\addtolength{\extrarowheight}{\belowrulesep}
\setlength{\aboverulesep}{0pt}
\setlength{\belowrulesep}{0pt}
\begin{tabular}{cc|cc} 
\toprule
\multicolumn{2}{c|}{Reward} & \multicolumn{2}{c}{Score} \\
\midrule
Rouge-L & EMS & Rouge-L & Exact Match \\ 
\hline
\greencheck  &  &{\cellcolor[rgb]{1,0.925,0.658}} \textbf{40.32} & 61.46 \\
& \greencheck & 24.06 & 61.75 \\
\greencheck & \greencheck &  39.69 & {\cellcolor[rgb]{1,0.925,0.658}}\textbf{69.80}  \\
\bottomrule
\end{tabular}
\caption{Models trained with different reward formulations on open-ended generation tasks.}
\label{tab: open-ended}
\end{table}

\noindent \textbf{Appropriate Reasoning Length.}
While test-time scaling methods have been shown to improve model performance in general-domain tasks such as mathematics and code generation, we observe that this trend does not consistently hold in the medical domain. Specifically, as shown in Figure~\ref{fig:token_acc}, increasing the reasoning length does not lead to significant performance gains. Most tasks reach a local optimum around a reasoning length of 512 tokens, and the overall performance curve resembles a bell-shaped (i.e., approximately normal) distribution. Interestingly, for simpler benchmarks such as MEDiQ, extending the reasoning length actually leads to a decrease in prediction accuracy. We hypothesize that these tasks rely more on intuitive prediction than on step-by-step reasoning, and that excessive reasoning may amplify consistency errors and hallucinations in model outputs.

\noindent \textbf{Impact of Reward Metric Selection.}
As described in Section~\ref{section: open-ended}, we explore three reward formulations for open-ended generation tasks: using only Rouge-L, using only EMS, and a Mix metric. As shown in Table~\ref{tab: open-ended}, the choice of reward function has a substantial impact on the quality of the final model outputs, highlighting the sensitivity of RL training to the reward design. the format reward encourages the model to generate concise and precise response within <answer></answer>, making EMS particularly well suited for tasks requiring structured outputs. However, relying solely on EMS as the reward signal tends to overfit the model to a small subset of short target labels, thereby impairing generalization performance on the test set. Thus, the Mix metric balances Rouge-L and EMS with equal weights, preserving competitive performance compared to using only Rouge-L, and achieving superior EMS scores, demonstrating improved output quality and better generalization.

\noindent \textbf{Emergence and Evolution of Thinking Patterns in Medical Tasks.}
By observing the training dynamics, we provide several insights into model adaptation and the emergence of reasoning. The response length initially increases slightly and then decreases, reflecting the transition of the model response from balancing $R_{format}$ and $R_{correct}$ to adopting a more faithful <think><answer> reasoning pattern. A subsequent rise and fall in length indicates a shift from exploratory, complex reasoning toward more stable and direct analytical prediction. Variations in the amplitude of response length fluctuations across tasks suggest differences in the distribution of effective reasoning signals within the training data for each task.

\noindent \textbf{Case Study of Different Training Paradigm.}
To better understand the nuanced differences in output behaviors across training paradigms (Base, SFT, and RL) and model scales, we conduct a qualitative case study. As illustrated in Figure~\ref{fig:case_study} (Appendix), after acquiring both background information about the patient and the specific clinical query, the model must identify which pieces of information are relevant to solve the problem and which are distractors. In this particular case, body weight serves as the key feature, whereas indicators like blood pressure and pulse are irrelevant to the final answer.
During the initial reasoning phase, all models can make a generally correct judgment. However, we observe that several models tend to second-guess the initial analysis. While this act of rethinking the initial reasoning is encouraging, it inadvertently leads to incorrect final outcomes in this case. We attribute this behavior to suboptimal sampling strategies and a lack of domain knowledge. In contrast, models trained with SFT or optimized through RL exhibit more stable and reliable reasoning paths, demonstrating improved robustness in their inferential process. Morever, in the following arithmetic reasoning, Med-U1 can arrive at correct answers more efficiently than models of comparable size, corroborating our experimental findings that reinforcement learning significantly enhances the model's reasoning and computational capabilities.

\noindent \textbf{Exploration of Open Generation Tasks Reward.}
Figure~\ref{fig:bleu} shows the training dynamics. As mentioned in Section~\ref{sec:Task-specific Reward} Failure Case, the model struggles to consistently achieve high reward $R_{correct}$, largely due to the rigid constraints imposed by the output format, highlighting a critical need for more flexible reward that better capture the semantic quality and correctness of generated answers.

\begin{figure}[t]
  \includegraphics[width=\columnwidth]{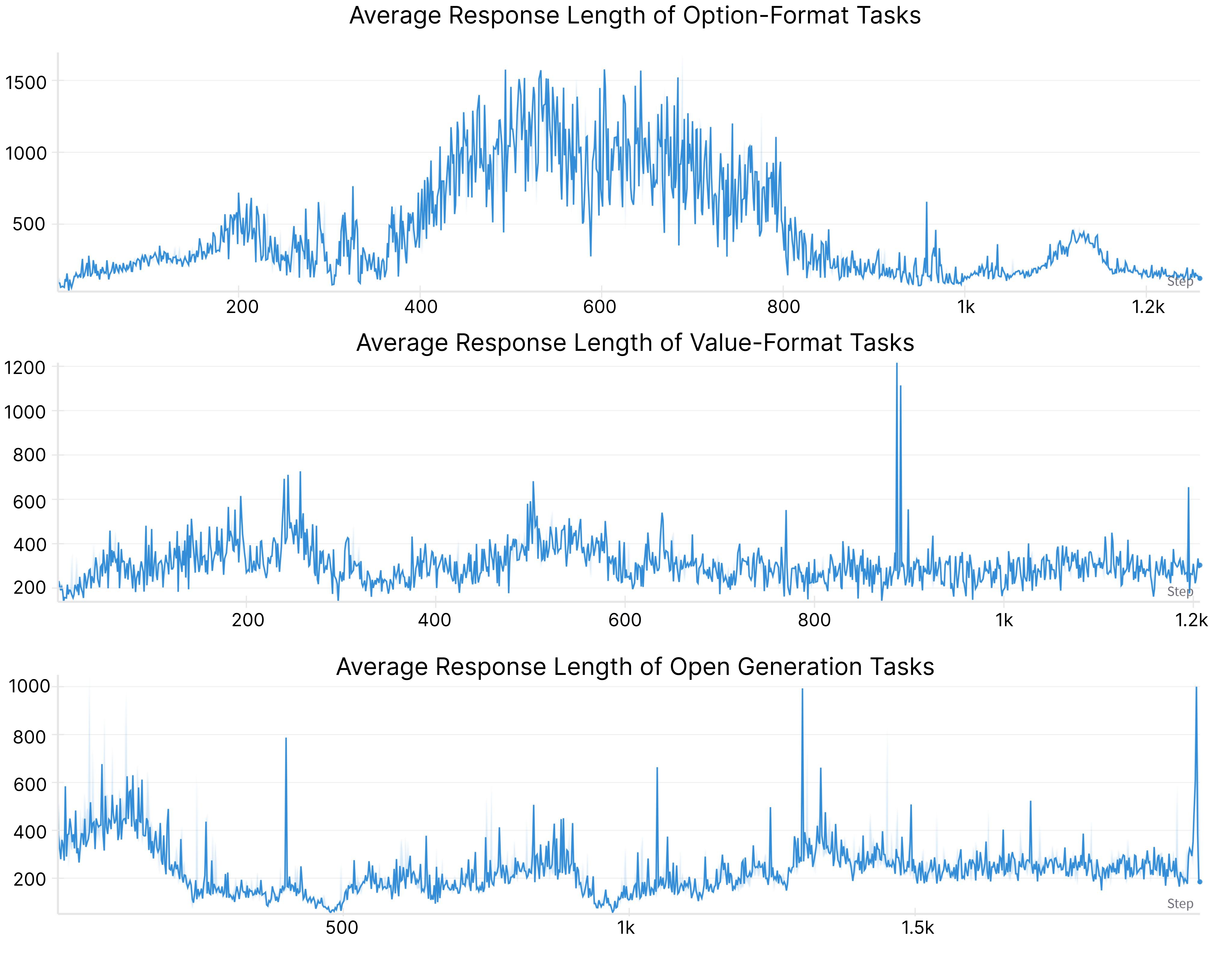}
  \caption{Average response length across various tasks over training steps. While the absolute lengths vary between tasks, the trends demonstrate a broadly consistent pattern of evolution during training.}
  \label{fig:response length}
\end{figure}

\section{Conclusion}

In this work, we introduce Med-U1, a unified and RL-based framework tailored for robust reasoning across diverse medical question-answering tasks. Empirical evaluations across multiple Med-QA benchmarks demonstrate that Med-U1 surpasses existing state-of-the-art models, and exhibits strong generalization capabilities to OOD tasks. We believe that Med-U1 lays a scalable foundation for future research on trustworthy and generalizable medical language models.

\section*{Limitations}

We acknowledge the presence of certain limitations. Although our findings suggest that there may exist an optimal reasoning length for medical tasks, our conclusion is constrained by computational resources. Specifically, we were only able to explore a limited range of reasoning lengths and thus can only identify a potential local optimum rather than a definitive global one. It remains uncertain whether model performance might recover or further improve at longer reasoning lengths beyond our current exploration. However, our results indicate that the model can already achieve competitive performance under relatively low inference budgets. Moreover, in open-ended scenarios, both the content and format of the model output tend to be significantly more complex. Although the reward design adopted in this work effectively addresses several issues we observed during training, the unique clinical applicability of such tasks suggests that further research is needed to develop reward functions specifically tailored to this setting. Designing more fine-grained and clinically grounded reward objectives remains an important direction for future work.

\newpage

\newpage
\appendix

\section*{Appendix}
\label{sec:appendix}

In this section, we present additional implementation details, experiment results, and supplements. The content structure is outlined as follows:

\begin{itemize}
    \item Section~\ref{appendix:related work} - Related Work 
    \begin{itemize}
        \item Section~\ref{Medical} - Medical Reasoning of LLMs
        \item Section~\ref{Length} - Length Control in LLMs
    \end{itemize}
    \item Section~\ref{Details} - Details of Experiments
    \begin{itemize}
        \item Section~\ref{appendix:dataset} - Dataset and Benchmarks
        \item Section~\ref{Description} - Description of medical-o1-reasoning SFT dataset
        \item Section~\ref{appendix:evaluation} - Evaluation Details
        \item Section~\ref{appendix:sft details} - SFT Training Details
        \item Section~\ref{appendix: task sensitivity} - Training Dynamics on Task Sensitivity Experiments
        \item Section~\ref{Training-BLEU} - Training Dynamics on BLEU Reward
    \end{itemize}
\end{itemize}

\section{Related Work}
\label{appendix:related work}
\subsection{Medical Reasoning of LLMs.} 
\label{Medical}
Previous medical LLMs have relied mainly on two approaches: task-specific prompting and further fine-tuning with medical data. However, the dependence on large-scale annotated datasets and the sensitiveness to task formulation significantly limit their generalizability across diverse medical subtasks~\cite{jiang2024medmoemixturedomainspecificexperts}. Although the o1 paradigm has demonstrated remarkable success in general domains, few studies have systematically investigated its application to deliberative chain-of-thought reasoning in clinical settings\citep{finemedlm}. HuatuoGPT-o1~\citep{chen2024huatuogpt} adopted a fine-tuning strategy that constructs complex reasoning trajectories, further enhanced through reinforcement learning. However, the process of generating reasoning paths and verifying answers is constrained by the capacity of the verifier and often incurs high computational costs. FineMedLM-o1~\citep{finemedlm} enhanced reasoning capabilities through multi-stage supervised fine-tuning and direct preference optimization, and further incorporates Test-Time Training (TTT) to promote domain adaptation and improve the reliability of reasoning. \citet{zhang2025med} explored the feasibility of Reinforcement Learning from Verifiable Rewards (RLVR) in the medical domain, highlighting the potential of pure reinforcement learning to improve model reliability and reasoning accuracy.

However, existing approaches predominantly focus on general medical scenarios, with the reward constructed in a multiple choice format. As a result, they lack a systematic exploration of how to design task-specific reward functions that generalize across heterogeneous medical tasks and supervision formats. In addition, the specific requirements of reasoning budget allocation, such as test-time inference length control, remain largely underexplored in clinical applications.

\subsection{Length Control in LLMs.} 
\label{Length}
Controlling the output length of LLMs is a critical practical consideration across various generation tasks. Prior efforts have explored different strategies to address this issue,~\citet{butcher2025precise} manipulated positional encodings to enforce precise sequence lengths;~\citet{yuan2024following} trained models on instruction-style data explicitly annotated with target output lengths; and~\citet{singhal2023long} modified training objectives to directly enforce length constraints. However, these methods have predominantly focused on general-purpose text generation or instruction-following contexts. The specific challenges of length control in medical text generation, where outputs often require both conciseness and completeness, remain underexplored and are not adequately addressed.

Moreover, the medical domain encompasses a wider range of complex scenarios and diverse subtasks, where the desired length of reasoning may vary substantially in different contexts. Despite this variability, length control tailored specifically for medical reasoning remains largely underexplored. Although certain works on general reasoning tasks have emphasized the generation of shorter reasoning chains to improve efficiency, these approaches often lack explicit mechanisms for fine-grained length control and cannot align precisely with user-specified reasoning budgets. In contrast, S1~\citep{muennighoff2025s1} introduced a strict token budget enforcement mechanism, referred to as "budget-forcing", which enables models to truncate outputs or insert a special token "Wait" to explicitly request continuation. Similarly,~\citet{aggarwal2025l1} proposed rule-based reward functions that penalize outputs that violate length constraints, enabling models to be trained with sensitivity to predefined length requirements.
\begin{figure*}[t]
  \includegraphics[width=\textwidth]{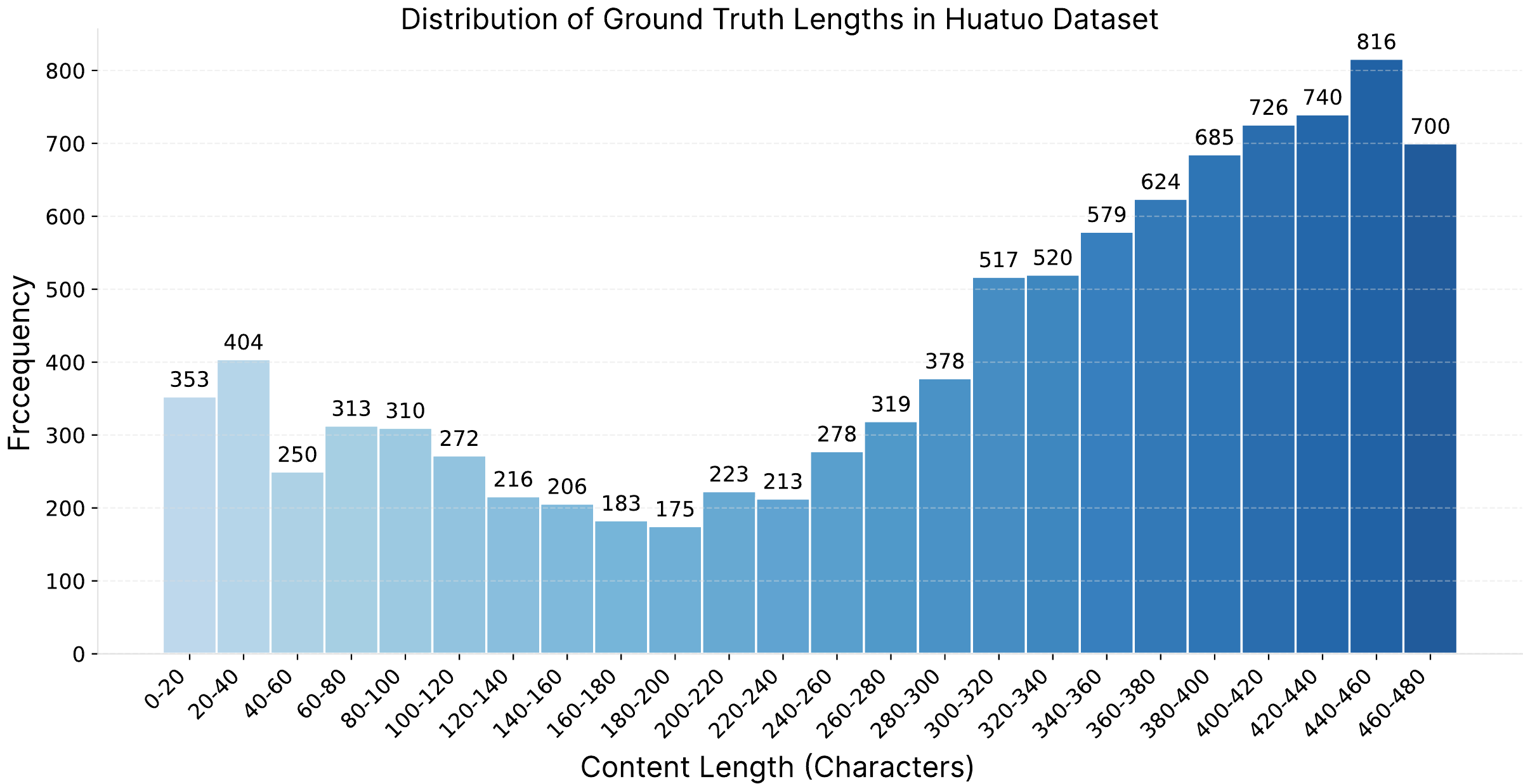}
  \caption{Character-level length distribution of all ground truth answers in the medical-o1-reasoning-SFT (Huatuo) dataset. Unlike value-based tasks that typically require only one or two words, open-ended generation involves free-form responses with varying lengths and formats. Notably, only 353 ground truth answers fall within the 0–20 character range, highlighting the general length variability of this dataset.}
  \label{fig:answer_length}
\end{figure*}

\section{Details of Experiments}
\label{Details}
\subsection{Datasets and Benchmarks}
\label{appendix:dataset}

\begin{figure*}[t]
  \includegraphics[width=\textwidth]{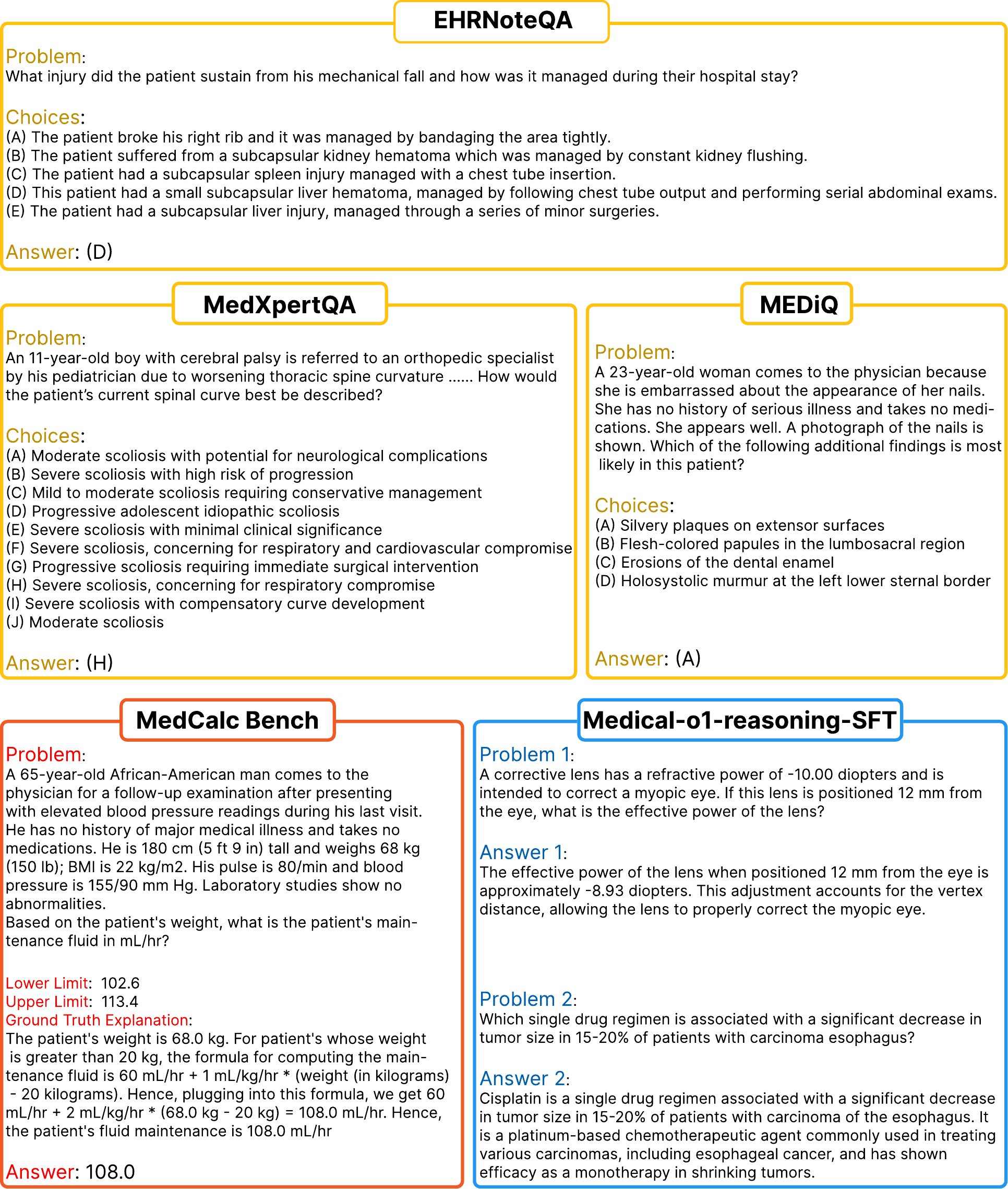}
  \caption{An overview of the five datasets used in our experiments is presented, with examples drawn from each. Datasets highlighted with yellow borders correspond to multiple choice tasks, those with red borders represent numerical reasoning tasks, and those with blue borders correspond to open-ended generation tasks. For each dataset, we provide the expected answer format as a reference, which informs the design of our rule-based reward function.}
  \label{fig:dataset-app}
\end{figure*}

To ensure comprehensive and reliable training and evaluation, we select a suite of recently introduced datasets that are well-suited for assessing long-form reasoning, with all sources explicitly documented. EHRNoteQA is constructed from the MIMIC-IV electronic health record (EHR) database and contains 962 question–answer pairs, each associated with a distinct patient’s discharge summary. EHRNoteQA includes questions that require information across multiple discharge summaries and covers ten diverse topics. MedXpertQA serves as a challenging and comprehensive benchmark designed to evaluate expert-level medical knowledge and advanced clinical reasoning skills. MEDiQ is originally developed to assess the question-asking capabilities of LLMs. However, due to the presence of rich contextual information that supports complex reasoning, we include it for the evaluation of reasoning abilities. MedCalc-Bench comprises more than 1,000 manually verified instances drawn from 55 distinct medical calculation tasks. It simulates real-world situations where physicians require quantitative reasoning and rule-based computations for evidence-based decision support. Medical-o1-reasoning-SFT is a synthetic dataset constructed using GPT-4o, where solutions to verifiable medical problems are automatically generated and validated by a medical verifier, enabling evaluation of multi-step reasoning and factual correctness. In Figure~\ref{fig:dataset}, we summarize the number of samples used from each dataset during training, along with the subset of cases involved in reward computation. The data sources are also annotated to ensure that all selected datasets are credible and clinically valuable. Detailed Data Examples can be seen in Figure ~\ref{fig:dataset-app}. 

\begin{figure}[htbp]
  \includegraphics[width=0.48\textwidth]{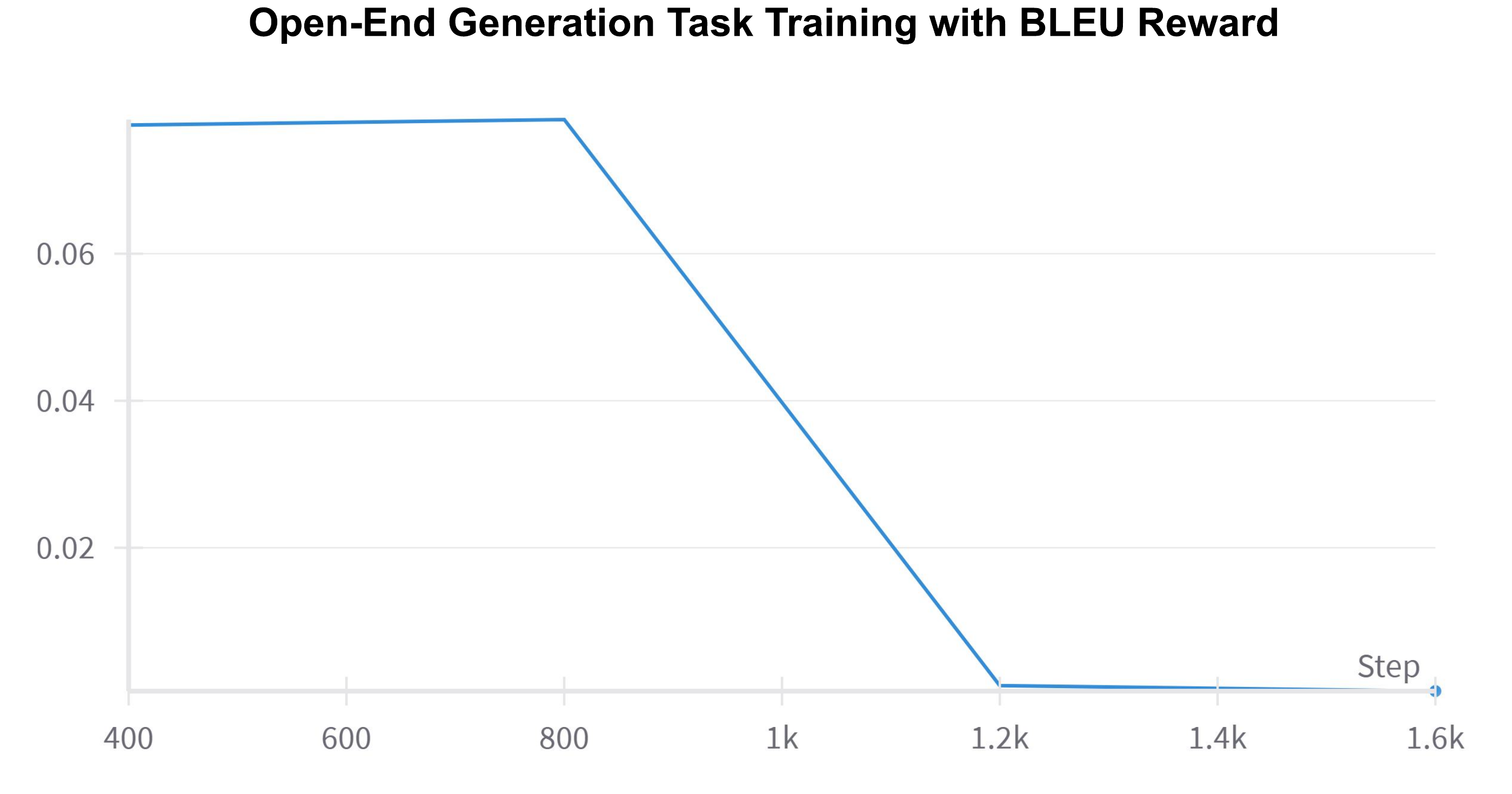}
  \caption{BLEU score as the reward signal.}
  \label{fig:bleu}
\end{figure}

\subsection{Description of medical-o1-reasoning SFT dataset.}
\label{Description}

In our text-format task, we have carefully considered the diversity of medical scenarios and the richness of medical answer formats. To achieve this, we utilized the medical-o1-reasoning-SFT dataset, which integrates the original MedQA-USMLE and MedMCQA datasets and transforms them into free-form answer formats. This transformation not only increases the difficulty of answering questions but also better reflects real-world medical problem-solving scenarios.

The medical-o1-reasoning-SFT dataset encompasses over 20 distinct medical disciplines, including Biochemistry, Anatomy, Physiology, and more. To further enhance the complexity of our task, we selected 10K challenging questions from this dataset. The answer length distribution of these questions, as shown in Figure~\ref{fig:answer_length}, is diverse, ranging from short phrases (under 5 characters) to long sentences (over 400 characters). Such a wide range of answer formats significantly increases the difficulty of our text-format task and highlights the advantages of using RL, which can better handle the complexity and variability of medical questions and answers, enabling the model to provide more accurate and diverse responses in different medical scenarios.

\subsection{Evaluation Details}
\label{appendix:evaluation}

When evaluating model performance on the test set, we deployed open-source models locally using frameworks such as vLLM\footnote{https://github.com/vllm-project/vllm} or HuggingFace\footnote{https://huggingface.co/docs/transformers/main\_classes/text\_generation} implementations. We use the sampling decoding strategy with a temperature of 0.2, and top\_p set to 0.95. The maximum generation length was capped at 2048 tokens.

\subsection{SFT Training Details}
\label{appendix:sft details}

For the Supervised Fine-Tuning (SFT) baseline shown in the main experiments (Section~\ref{section: in domain}) and the task sensitivity analysis (Section~\ref{section: task sensitivity}), we utilize the LLaMA-Factory~\citep{zheng2024llamafactory} for training. For the former, we use the full set of training data described in Section~\ref{section: datasets}. The model is trained on NVIDIA H200 140G GPU for 8 epochs with a learning rate of e-5 and a batch size of 1. For the latter, which is used as a cold start initialization for the subsequent RL phase, we fine-tune the model only on the medical-o1-reasoning-SFT dataset annotated with <think></think> and <answer></answer> tags. This model is trained for 3 epoch with the same learning rate and batch size.

\subsection{Training Dynamics on Task Sensitivity Experiments.}
\label{appendix: task sensitivity}
\begin{figure*}[t]
  \includegraphics[width=0.98\textwidth]{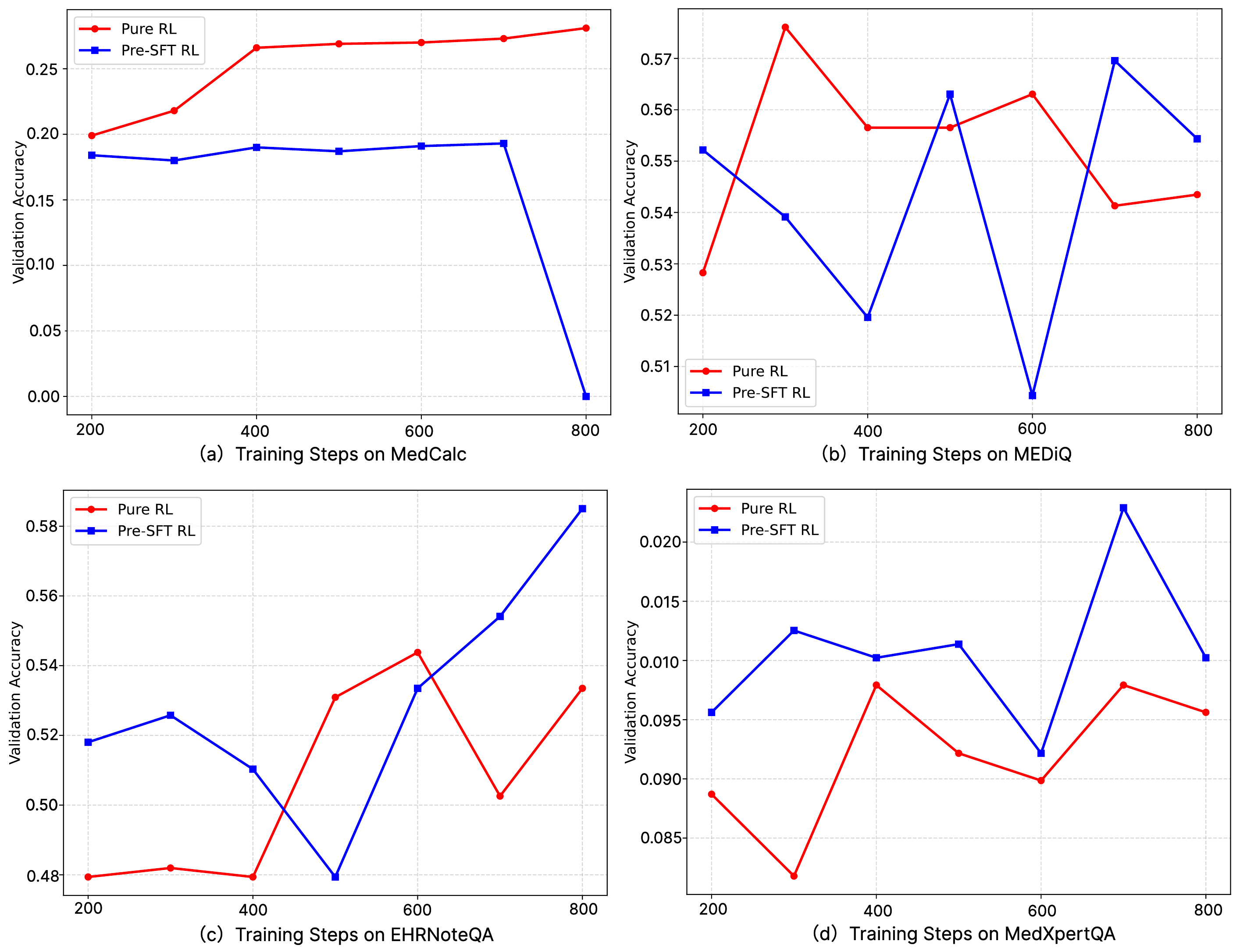}
  \caption{Training dynamics under different training strategies. The red curve denotes the pure RL paradigm, while the blue curve represents the SFT-then-RL approach.}
  \label{fig:training dynamic}
\end{figure*}

As shown in Figure~\ref{fig:training dynamic} (a), training with pure reinforcement learning in the MedCalc dataset leads to a more stable validation accuracy and improved performance relative to the pre-SFT-then-RL paradigm. Since MedCalc-Bench involves numerical reasoning tasks in a medical context, it requires models to possess strong computational reasoning abilities. Although SFT enriches the model with medical domain knowledge, it may interfere with the underlying general reasoning capabilities~\citep{huang2025vision,abdin2025phi}. Figure~\ref{fig:training dynamic} (b) and (c) further validate this trend: SFT provides a strong initialization by enhancing the model’s understanding of domain-specific terms, whereas training solely with RL leads to more stable and gradually improving performance. Thus, for tasks that involve complex numerical inference in medicine, we find pure RL to be a more appropriate training strategy.

\subsection{Training Dynamics on BLEU Reward}
\label{Training-BLEU}
When using BLEU as the reward signal to guide training on open-ended generation tasks, we observe catastrophic training collapse once the model fails to discover trajectories yielding higher reward.

\begin{figure*}[t]
  \includegraphics[width=\textwidth]{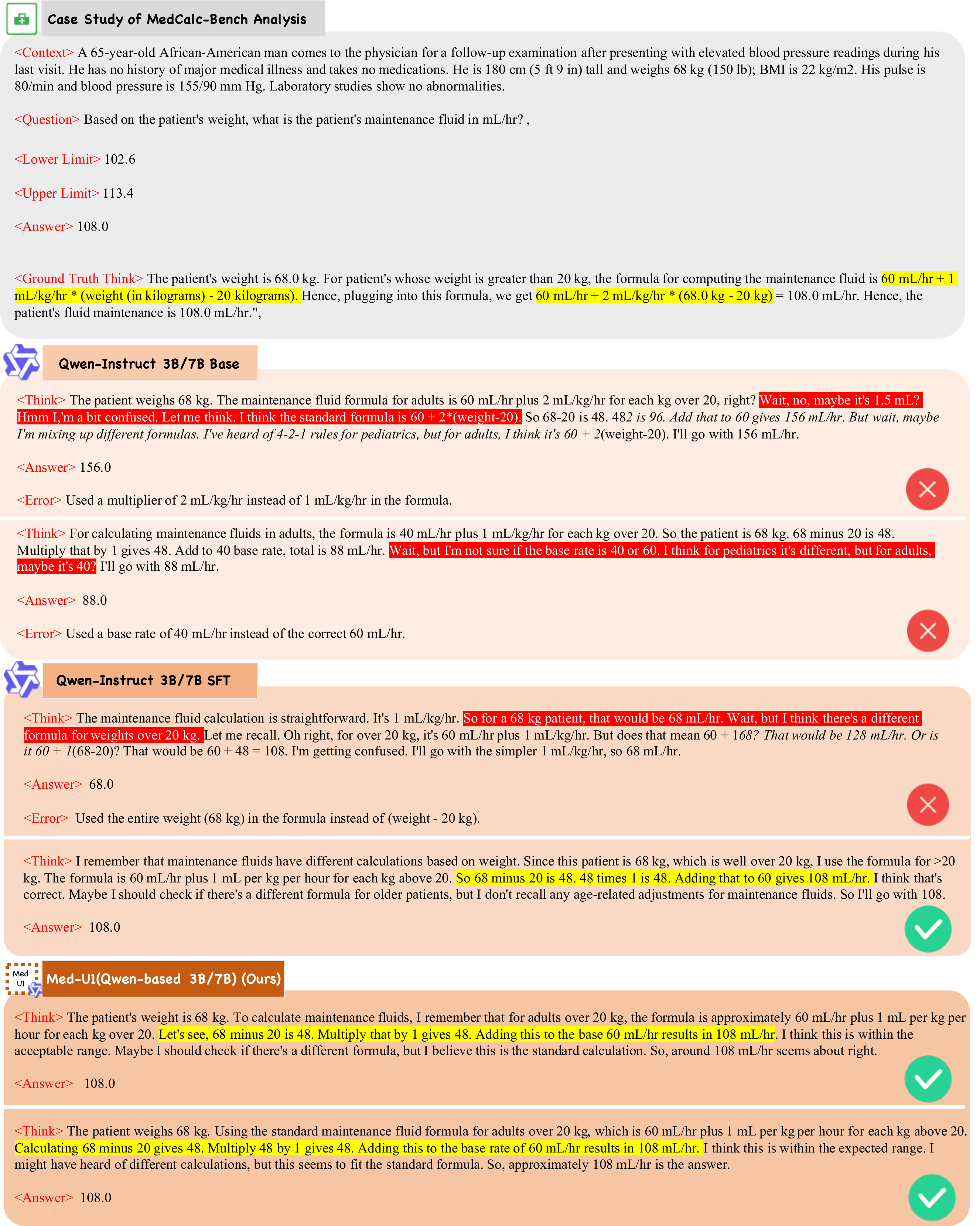}
  \caption{A case study comparing Qwen-Instruct-Base, Qwen-Instruct-SFT, and our Med-U1 method on MedCalc-Bench, highlighting how Med-U1 effectively addresses model hallucination and calculation inaccuracies in determining patient maintenance fluid requirements.}
  \label{fig:case_study}
\end{figure*}

\end{document}